\documentclass[sigconf]{acmart}

\usepackage{balance}
\usepackage{booktabs}
\usepackage{graphicx}
\usepackage{tikz}
\usepackage{subcaption}
\usepackage[table]{xcolor}
\usepackage{multirow}
\usepackage{makecell}
\usepackage{tabularx}
\usepackage[ruled,vlined]{algorithm2e}
\usepackage{enumitem}

\usetikzlibrary{positioning, arrows.meta, shapes.geometric, fit, calc, backgrounds}
\copyrightyear{2026}
\acmYear{2026}
\setcopyright{cc}
\setcctype{by}
\acmConference[ICCAD '26]{IEEE/ACM International Conference on Computer-Aided Design}{November 08--12, 2026}{San Jose, CA, USA}
\acmBooktitle{IEEE/ACM International Conference on Computer-Aided Design (ICCAD '26), November 08--12, 2026, San Jose, CA, USA}
\acmDOI{10.1145/3831252.3834249}
\acmISBN{979-8-4007-2873-0/2026/11}

\begin{document}

\title{Mesh-Native Physics-Informed Graph Surrogates for TCAD-in-the-Loop Design Space Exploration}

\author{Leonid Popryho}
\authornote{Corresponding author.}
\orcid{0009-0002-0578-9592}
\email{lpopry2@uic.edu}
\affiliation{%
    \institution{University of Illinois Chicago}
    \city{Chicago}
    \state{IL}
    \country{USA}
}

\author{Ayoub Sadeghi}
\orcid{0000-0001-9904-9813}
\email{asadeg3@uic.edu}
\affiliation{%
    \institution{University of Illinois Chicago}
    \city{Chicago}
    \state{IL}
    \country{USA}
}

\author{Inna Partin-Vaisband}
\orcid{0000-0002-6399-6672}
\email{vaisband@uic.edu}
\affiliation{%
    \institution{University of Illinois Chicago}
    \city{Chicago}
    \state{IL}
    \country{USA}
}

\begin{abstract}
    High-fidelity technology computer-aided design (TCAD) simulation of drift-diffusion transport remains the workhorse of emerging FinFET device design, but it is computationally expensive, particularly for three-dimensional (3D) structures, where runtime escalates steeply with mesh complexity. This cost sharply limits multi-objective design space exploration (DSE). Existing machine-learning surrogates map a fixed set of design parameters to a few scalar device metrics, discarding the underlying physics and losing transferability across device geometries and device families.

    A physics-informed graph attention network (GAT) surrogate is proposed. It operates directly on the tetrahedral TCAD mesh and predicts, at every mesh node, the electrostatic potential together with the electron and hole quasi-Fermi levels---the fundamental unknowns of the drift-diffusion system. Training combines a data loss with finite-volume current-continuity residuals, embedding carrier-transport physics directly into the objective.

    Operating on the mesh as a graph, the surrogate inherits size generalization: a model trained on few-fin meshes applies unchanged to substantially larger arrays, bounded at inference only by GPU memory. Per-node uncertainty from a deep ensemble drives an active-learning loop that screens large candidate pools in seconds and forwards only the most informative designs for full simulation.

    Benchmarked against Sentaurus Device on multi-fin tri-gate FinFETs, the surrogate reproduces the three drift-diffusion fields with a sub-volt per-field RMSE against the ground-truth TCAD result and, with GPU-accelerated inference, reaches a per-design throughput \emph{orders of magnitude higher than the full simulator}. The advantage becomes more significant with device size: on large multi-fin arrays that are prohibitively slow to simulate directly, inference still completes in under a second per device, enabling Pareto-front exploration across device scales that remain infeasible for direct TCAD sweeps.
\end{abstract}



\maketitle

\section{Introduction}
\label{sec:introduction}
This paper presents a \emph{general, mesh-native, physics-informed graph-attention surrogate} for TCAD-driven design space exploration, demonstrated on multi-fin tri-gate GaN FinFETs for vertical power delivery as a representative test case.
Design of modern semiconductor devices for advanced computing is constrained by the cost of accurate exploration across high-dimensional design spaces.
Roadmaps such as the IEEE Heterogeneous Integration Roadmap (HIR) emphasize aggressive 2.5D/3D integration, further tightening performance and efficiency requirements~\cite{chen_vlsi_2019}.
In vertical power delivery (VPD) architectures, power switches must meet strict conversion ratios ($48\,\mathrm{V}$-to-$1\,\mathrm{V}$), current density targets exceeding $2\,\mathrm{A/mm}^2$, and end-to-end efficiencies approaching 90\%~\cite{krishnakumar_ectc_2024}.
Multi-fin tri-gate gallium nitride (GaN) FinFETs, alongside other vertically integrated GaN power devices, have emerged as leading candidates for the VPD power stage~\cite{sadeghi_ectc_2025,wu_ted_2019,jeong_iedm_2024,nela_natel_2021}, but their performance depends on many structural parameters that interact nonlinearly, raising the cost of design-space exploration.

The increasing complexity of these emerging devices precludes accurate analytical models for rapid performance evaluation. TCAD offers high-fidelity device simulation, but each 3D run requires assembling and solving a large system on a tetrahedral mesh (i.e., a 3D discretization into irregular volumetric elements) whose cost grows steeply with mesh size: a single design point can take minutes to hours on a modern workstation~\cite{synopsys_manual_2022}.
Three factors compound to make TCAD-driven exploration prohibitive: \textit{(i)}~the design space is low-dimensional but highly nonconvex and multi-modal, requiring dense sampling; \textit{(ii)}~the optimization is inherently multi-objective, targeting a Pareto front rather than a single optimum; \textit{(iii)}~at minutes to hours per evaluation, resolving this front exceeds typical project and license budgets. Classical black-box optimizers---grid search, NSGA-II~\cite{deb_tevc_2002}, and manual expert tuning~\cite{sadeghi_ectc_2025,yu_tnano_2020}---cannot lift these constraints, because every query still requires a full TCAD solve.

Machine-learning surrogates have been proposed to amortize this cost, typically by regressing a fixed vector of design parameters onto a small set of scalar device figures of merit (FoM) such as drive current $I_{d,\max}$, threshold voltage $V_\mathrm{th}$, on-resistance $R_\mathrm{on}$, transconductance $g_m$, subthreshold swing, drain-induced barrier lowering (DIBL), and breakdown voltage~\cite{mehta_edl_2021,ko_ted_2019,akbar_ted_2021,rawat_acsomega_2022}.
These parameter-to-scalar maps are lightweight and fast to train, but they discard the underlying physics: the electrostatic and carrier fields are reduced to a few aggregates, the model has no notion of the device geometry beyond the input vector, and a surrogate trained on one mesh topology cannot transfer to a structurally different device without full retraining. Recent work that learns directly on device meshes~\cite{jang_edl_2023} or leverages physics-informed inductive biases~\cite{raissi_jcp_2019,li_iclr_2021,kim_sispad_2020} has shown that geometry-aware and physics-informed approaches recover fidelity and transferability, yet these strands have not been combined on full 3D drift-diffusion surrogates inside an active-learning loop for TCAD-in-the-loop design space exploration (DSE); Sec.~\ref{sec:related_work} surveys these threads in detail.

A physics-informed graph attention network (GAT) surrogate is proposed that operates directly on the tetrahedral TCAD mesh (i.e., the discretized device geometry), predicting the three drift-diffusion field unknowns---the electrostatic potential and the electron and hole quasi-Fermi levels---at every mesh node. Operating on the mesh rather than on a fixed parameter vector \textit{(i)}~aligns the surrogate with the finite-volume scheme used to enforce current continuity during training, \textit{(ii)}~inherits the size-invariance of graph neural networks so that a model trained on few-fin meshes transfers unchanged to larger arrays, and \textit{(iii)}~decouples the input from any specific device layout. Training augments a data loss with finite-volume current-continuity residuals evaluated on the mesh edges, embedding carrier-transport physics into the objective~\cite{raissi_jcp_2019} rather than applying it as a post-hoc correction. At inference time, disagreement across a deep ensemble yields a per-node uncertainty signal that drives the active-learning outer loop: large candidate pools are screened in seconds and only the most informative designs are forwarded to full Sentaurus simulation~\cite{lakshminarayanan_neurips_2017}. Benchmarked on multi-fin tri-gate GaN FinFETs, the surrogate reproduces the drift-diffusion fields with sub-volt RMSE per field, while the speed advantage over direct TCAD increases with device size. This enables Pareto-front exploration across device scales that are otherwise infeasible with direct TCAD sweeps. The mesh-graph abstraction, finite-volume physics loss, and uncertainty-driven active-learning loop are device-agnostic; the GaN FinFET study here uses a material-specific node embedding (GaN, AlGaN, HfO\textsubscript{2}, TiN, Au), and extending to other device families through a general learned material embedding is left to future work (Sec.~\ref{sec:conclusion}).
%
The contributions of this work are as follows:
\begin{itemize}[leftmargin=*, itemsep=0pt, topsep=1pt]
    \item \textbf{Discretization-consistent, mesh-native drift-diffusion surrogate.}
          A graph attention network is formulated directly on the tetrahedral TCAD mesh, predicting the three fundamental drift-diffusion unknowns---the electrostatic potential and the electron and hole quasi-Fermi levels---at every mesh node. Unlike prior parameter-to-scalar surrogates or mesh-based predictors without full-field reconstruction, this formulation aligns learning with the TCAD discretization and preserves device physics.

    \item \textbf{Physics-informed training aligned with the finite-volume solver.}
          Training integrates a data loss with finite-volume current-continuity residuals computed on the same mesh edges used by the TCAD solver. This embeds carrier-transport physics at the discretization level as a soft regularizer, avoiding the limitations of continuous physics-informed neural networks at material interfaces and keeping the surrogate aligned with the ground-truth numerical scheme.

    \item \textbf{Size-transferable surrogate across device geometries.}
          By operating on mesh graphs with localized message passing, the surrogate generalizes across device sizes, enabling models trained on few-fin structures to transfer without retraining to substantially larger multi-fin arrays. This opens efficient exploration of device scales that are prohibitively expensive for direct TCAD sweeps.

    \item \textbf{Field-aware uncertainty for TCAD-in-the-loop design space exploration.}
          A deep ensemble produces per-node uncertainty estimates, enabling spatially resolved confidence assessment of predicted fields. This uncertainty drives an active-learning loop that prioritizes informative designs for TCAD evaluation, enabling surrogate-assisted exploration while maintaining high-fidelity simulation where needed. The entire training set used in this paper was itself assembled through this loop (Sec.~\ref{sec:experiments}): 147 bootstrap designs, expanded by 53 acquisition-selected designs concentrated on the least-certain device sizes.
\end{itemize}


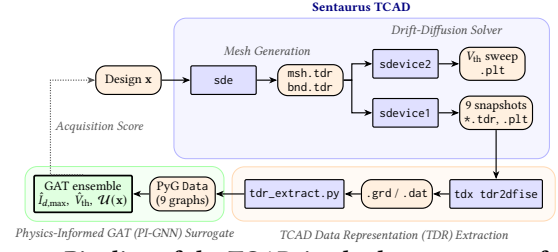
\begin{figure}[!t]
    \centering
    \resizebox{0.85\linewidth}{!}{%
        \begin{tikzpicture}[
            font=\normalsize,
            node distance=3mm and 5mm,
            >={Stealth[length=2mm, width=1.6mm]},
            exe/.style={
                    draw, thick, rectangle, rounded corners=1pt,
                    minimum height=7mm, minimum width=16mm, align=center,
                    fill=blue!12,
                },
            data/.style={
                    draw, thick, rectangle, rounded corners=2.5mm,
                    minimum height=7mm, minimum width=16mm, align=center,
                    fill=orange!12,
                },
            model/.style={
                    draw, very thick, rectangle, rounded corners=1pt,
                    minimum height=9mm, minimum width=24mm, align=center,
                    fill=green!12,
                },
            arr/.style={->, thick},
            reuse/.style={->, thick, dashed, densely dotted, gray!70!black},
            lbl/.style={font=\itshape},
            stagelbl/.style={font=\itshape, text=black!70},
            ]

            \node[data] (x) {Design $\mathbf{x}$};

            \node[exe, right=7mm of x] (sde) {\texttt{sde}};
            \node[data, right=5mm of sde] (tdr1)
            {\texttt{msh.tdr}\\[-0.5mm]\texttt{bnd.tdr}};

            \node[exe, right=7mm of tdr1, yshift=-8mm] (sd1) {\texttt{sdevice1}};
            \node[exe, right=7mm of tdr1, yshift=4mm]  (sd2) {\texttt{sdevice2}};

            \node[data, right=5mm of sd2]
            (snap2) {$V_{\mathrm{th}}$ sweep\\[-0.5mm]\texttt{.plt}};
            \node[data, right=5mm of sd1]
            (snap1) {9 snapshots\\[-0.5mm]\texttt{*.tdr}, \texttt{.plt}};

            \node[exe, below=12mm of snap1] (tdx)
            {\texttt{tdx tdr2dfise}};
            \node[data, left=4mm of tdx]
            (dfise) {\texttt{.grd} / \texttt{.dat}};
            \node[exe, left=4mm of dfise]
            (extract) {\texttt{tdr\_extract.py}};

            \node[data, left=7mm of extract] (pyg)
            {PyG \texttt{Data}\\[-0.5mm](9 graphs)};

            \node[model, left=4mm of pyg] (gat)
            {GAT ensemble\\[-0.5mm]$\hat I_{d,\max},\ \hat V_{\mathrm{th}},\ \mathcal{U}(\mathbf{x})$};

            \draw[arr] (x) -- (sde);
            \draw[arr] (sde) -- (tdr1);
            \draw[arr] (tdr1.east) -- ++(2mm,0) |- (sd1.west);
            \draw[arr] (tdr1.east) -- ++(2mm,0) |- (sd2.west);
            \draw[arr] (sd1) -- (snap1);
            \draw[arr] (sd2) -- (snap2);

            \draw[arr] (snap1.south) -- (tdx.north);

            \draw[arr] (tdx) -- (dfise);
            \draw[arr] (dfise) -- (extract);
            \draw[arr] (extract) -- (pyg);
            \draw[arr] (pyg) -- (gat);

            \draw[reuse]
                (gat.north) -- ++(-8mm,0)
                |- node[pos=0.25, right, lbl] {Acquisition Score}
                (x.west);

            \begin{scope}[on background layer]
                \node[draw=blue!30, fill=blue!4, rounded corners=3mm,
                    inner xsep=4mm, inner ysep=7mm,
                    fit=(sde)(tdr1)(sd1)(sd2)(snap1)(snap2)] (tcad_box) {};

                \node[draw=orange!30, fill=orange!4, rounded corners=3mm,
                    inner sep=3mm,
                    fit=(tdx)(dfise)(extract)] (conv_box) {};

                \node[draw=green!35, fill=green!4, rounded corners=3mm,
                    inner sep=2mm,
                    fit=(pyg)(gat)] (surr_box) {};
            \end{scope}

            \node[font=\normalsize\bfseries, text=blue!50!black]
            at (tcad_box.north) [above=0.5mm] {Sentaurus TCAD};

            \node[stagelbl]
            at ($(sde.north)!0.5!(tdr1.north) + (0,3.5mm)$)
            {Mesh Generation};

            \node[stagelbl]
            at ($(sd1.south)!0.5!(sd2.north) + (10mm,14mm)$)
            {Drift-Diffusion Solver};

            \node[stagelbl]
            at (conv_box.south) [below=1mm]
            {TCAD Data Representation (TDR) Extraction};

            \node[stagelbl]
            at (surr_box.south) [below=0.5mm]
            {Physics-Informed GAT (PI-GNN) Surrogate};

        \end{tikzpicture}%
    }
    \vspace{-12pt}
    \caption{Pipeline of the TCAD-in-the-loop surrogate framework. A parameterized device design $\mathbf{x}$ is converted to a tetrahedral mesh (sde) and evaluated by the drift-diffusion solver (SDevice) to produce bias-dependent field snapshots. These are converted into mesh-aligned graphs and used to train a physics-informed GAT ensemble that predicts per-node fields and scalar objectives. The dashed path denotes the active-learning loop, where uncertainty-driven acquisition selects new designs for full TCAD evaluation.}
    \label{fig:framework}
    \Description{Two-row block diagram of the TCAD-in-the-loop active learning workflow. The top row shows the Sentaurus pipeline from design parameters through mesh generation and drift-diffusion simulation. The bottom row shows TDR extraction feeding into the PI-GNN surrogate. A dashed arrow on the left indicates the active-learning feedback loop from the surrogate back to the design input.}
    \vspace{-10pt}
\end{figure}

\section{Related Work}
\label{sec:related_work}

\begin{table}[!t]
\centering
\caption{Representative ML surrogates for TCAD/PDE. 
Rep: M=mesh-native, G=grid. 
Phys: PDE=residual, FV=finite-volume, out=output constraint.}
\label{tab:related-work}
\vspace{-8pt}
\footnotesize
\setlength{\tabcolsep}{1.3pt}
\newcommand{\tightcell}[2]{\shortstack[c]{#1\\[-3pt]#2}}
\renewcommand{\arraystretch}{1.15}
\begin{tabular}{@{}p{2.3cm}p{1.3cm}cccccc@{}}
\toprule
\multirow{2}{*}{Category} & \multirow{2}{*}{Refs}
& \multicolumn{2}{c}{Represent.}
& \multicolumn{2}{c}{Physical loss}
& \multirow{2}{*}{Size}
& \multirow{2}{*}{Uncert.} \\
\cmidrule(lr){3-4} \cmidrule(lr){5-6}
& & Mesh & Field & Type & Level & & \\
\midrule
Scalar-FoM surrogate 
& \cite{mehta_edl_2021,ko_ted_2019,akbar_ted_2021,rawat_acsomega_2022}
& $\times$ & $\times$ & $\times$ & $\times$ & $\times$ & $\times$ \\

Warm-start regressor
& \cite{han_ted_2021}
& $\times$ & $\checkmark$ & $\times$ & $\times$ & $\times$ & $\times$ \\

Physics-guided NCM
& \cite{kim_sispad_2020}
& $\times$ & $\times$
& \tightcell{output}{constr.}
& \tightcell{FoM}{post-proc.}
& $\times$ & $\times$ \\

Mesh GNN (TCAD) 
& \cite{jang_edl_2023}
& $\checkmark$ & $\checkmark$ & $\times$ & $\times$ & $\times$ & $\times$ \\

Mesh GNN (PDE) 
& \cite{pfaff_iclr_2021,brandstetter_iclr_2022}
& $\checkmark$ & $\checkmark$ & $\times$ & $\times$ & $\checkmark$ & $\times$ \\

Neural operator
& \cite{li_iclr_2021}
& \tightcell{grid}{only} & $\checkmark$ & $\times$ & $\times$ & $\checkmark$ & $\times$ \\

PINN (continuous)
& \cite{raissi_jcp_2019}
& $\times$ & $\checkmark$
& \tightcell{PDE}{residual}
& \tightcell{autograd}{colloc.}
& $\times$ & $\times$ \\

AL / multi-obj. DSE
& \cite{deb_tevc_2002,shahriari_pieee_2016,daulton_neurips_2020,park_dac_2024}
& $\times$ & $\times$ & $\times$ & $\times$ & N/A & \tightcell{scalar}{per-trial} \\

\midrule
\textbf{This work}
& --
& $\checkmark$ & $\checkmark$
& \tightcell{PDE}{residual}
& \tightcell{FV/mesh}{edges}
& $\checkmark$ & ensemble \\
\bottomrule
\end{tabular}
\vspace{-8pt}
\end{table}

Table~\ref{tab:related-work} positions each line of prior work relative to the capabilities of the proposed surrogate.

\noindent\textbf{\textit{Scalar-FoM surrogates for TCAD.}}
Early ML surrogates for TCAD map a fixed vector of device and process parameters to scalar FoMs or to I--V/C--V curves sampled at a small set of applied gate and drain voltages. Autoencoder-based compact models reproduce full FinFET I--V/C--V responses from modest amounts of TCAD data and report high agreement on $I_\mathrm{ON}$, $I_\mathrm{OFF}$, DIBL and $g_m$~\cite{mehta_edl_2021}. Neural networks have also been used as warm-start predictors, supplying Newton initial guesses that shorten per-bias drift-diffusion solves without altering the underlying TCAD engine~\cite{han_ted_2021}. In variability-heavy settings, such as work-function fluctuation in nanosheet gate-all-around devices or process-variation effects in advanced nodes, surrogate regressors avoid otherwise intractable 3D Monte Carlo sweeps~\cite{akbar_ted_2021,ko_ted_2019}, and meta-learned, TCAD-assisted sampling further improves data efficiency across related process conditions~\cite{rawat_acsomega_2022}. These approaches share a common limitation: the field solution is reduced to a few scalars, the model has no notion of the geometry beyond the input vector, and transfer to a structurally different device requires a new dataset and retraining from scratch.

\noindent\textbf{\textit{Learning on meshes and neural PDE solvers.}}
Graph neural networks that pass messages on local mesh neighborhoods have emerged as a natural fit for unstructured simulation data. Mesh graph networks establish that learned local update rules on irregular meshes generalize to domains and resolutions unseen during training~\cite{pfaff_iclr_2021}, and neural partial differential equation (PDE) solvers extend the approach to time-dependent problems with explicit stability arguments~\cite{brandstetter_iclr_2022}. Within device simulation, graph neural networks applied directly to geometry and doping predict TCAD-quality outputs from small datasets~\cite{jang_edl_2023}, demonstrating mesh-native learning on modest single-device scales but without a physics-aware training signal. 
The surrogate combines unstructured-mesh message passing with a finite-volume physics loss, a combination not previously exercised on full 3D drift-diffusion.

\noindent\textbf{\textit{Physics-informed learning for transport.}}
Physics-informed neural networks embed governing PDEs as a soft penalty, using automatic differentiation to evaluate residuals at randomly sampled collocation points in the continuous domain~\cite{raissi_jcp_2019}. This approach presumes that the predicted field is smooth enough for gradients and Laplacians computed by autograd (automatic differentiation) to remain meaningful, an assumption that breaks down at abrupt material junctions where the electrostatic potential is continuous but its derivative jumps by orders of magnitude across a heterointerface. Physics-augmented neural compact models instead encode device laws as monotonicity or scaling constraints on post-processed figures of merit rather than as PDE residuals~\cite{kim_sispad_2020}, which preserves physical consistency at the output level but does not constrain the underlying field. The proposed surrogate evaluates current-continuity residuals directly on the same finite-volume scheme used by Sentaurus Device, enforcing conservation at the discretization level rather than at autograd collocation points, masking contact nodes from the residual since current enters and exits them.

\noindent\textbf{\textit{Active learning and multi-objective DSE over expensive solvers.}}
Evolutionary search such as NSGA-II remains a default EDA baseline for black-box multi-objective exploration but treats every query as a full device simulation~\cite{deb_tevc_2002}. Bayesian optimization improves sample efficiency by coupling a Gaussian process surrogate with an acquisition function~\cite{shahriari_pieee_2016,garnett_book_2023}; multi-objective variants such as ParEGO and (q)EHVI target Pareto-front progress~\cite{knowles_tevc_2006,daulton_neurips_2020}, while best-practice surveys caution that Gaussian process effectiveness degrades in high-dimensional, heteroscedastic, constraint-heavy regimes typical of TCAD~\cite{siivola_aail_2021}. Multi-output Gaussian processes share representations across correlated objectives~\cite{bonilla_neurips_2007}, while deep ensembles provide robust epistemic uncertainty estimates for neural surrogates and are naturally parallelizable~\cite{lakshminarayanan_neurips_2017}. Recent device-level design-technology co-optimization (DTCO) flows wrap TCAD in uncertainty-guided active-learning loops and report meaningful wall-clock savings over random search~\cite{park_dac_2024}. All of these approaches operate on a \emph{scalar-per-trial} uncertainty signal tied to a few output metrics. The proposed surrogate instead produces a \emph{per-node} uncertainty field from ensemble disagreement, so candidate devices are screened at the field level---revealing where in the device the ensemble disagrees---before a full TCAD run is committed.

\section{Method}
\label{sec:method}
This section presents the mesh-native physics-informed graph attention (GAT) surrogate and the active-learning loop in which it is embedded, in four parts: problem formulation, mesh-graph representation, physics-informed GAT surrogate, and TCAD-in-the-loop active learning. The overall workflow is sketched in Fig.~\ref{fig:framework}.

\begin{figure*}[!ht]
    \centering
    \vspace{-12pt}
    \begin{minipage}[c]{0.32\textwidth}
        \centering
        \refstepcounter{table}\label{tab:ranges}
\footnotesize
\renewcommand{\arraystretch}{1.15}
\setlength{\tabcolsep}{2pt}
\begin{tabularx}{\linewidth}{@{}>{\raggedright\arraybackslash}X c c c@{}}
    \toprule
    Parameter & Symbol & Baseline ~\cite{sadeghi_ectc_2025} & Range \\
    \midrule
    GaN thickness (nm)       & $T_{\mathrm{GaN}}$   & 20        & 10--100 \\
    AlGaN thickness (nm)     & $T_{\mathrm{AlGaN}}$ & 40        & 10--100 \\
    Fin width (nm)           & $W_{\mathrm{fin}}$   & 100       & 10--250 \\
    Oxide thickness (nm)     & $T_{\mathrm{ox}}$    & 6         & 5--25 \\
    Gate thickness (nm)      & $T_{\mathrm{Gate}}$  & 20        & 5--40 \\
    AlGaN doping (cm$^{-3}$) & $N_{\mathrm{AlGaN}}$ & $10^{18}$ & $10^{18}$--$10^{20}$ \\
    \bottomrule
\end{tabularx}
    \end{minipage}\hfill
    \begin{minipage}[c]{0.24\textwidth}
        \centering
        \includegraphics[width=\linewidth]{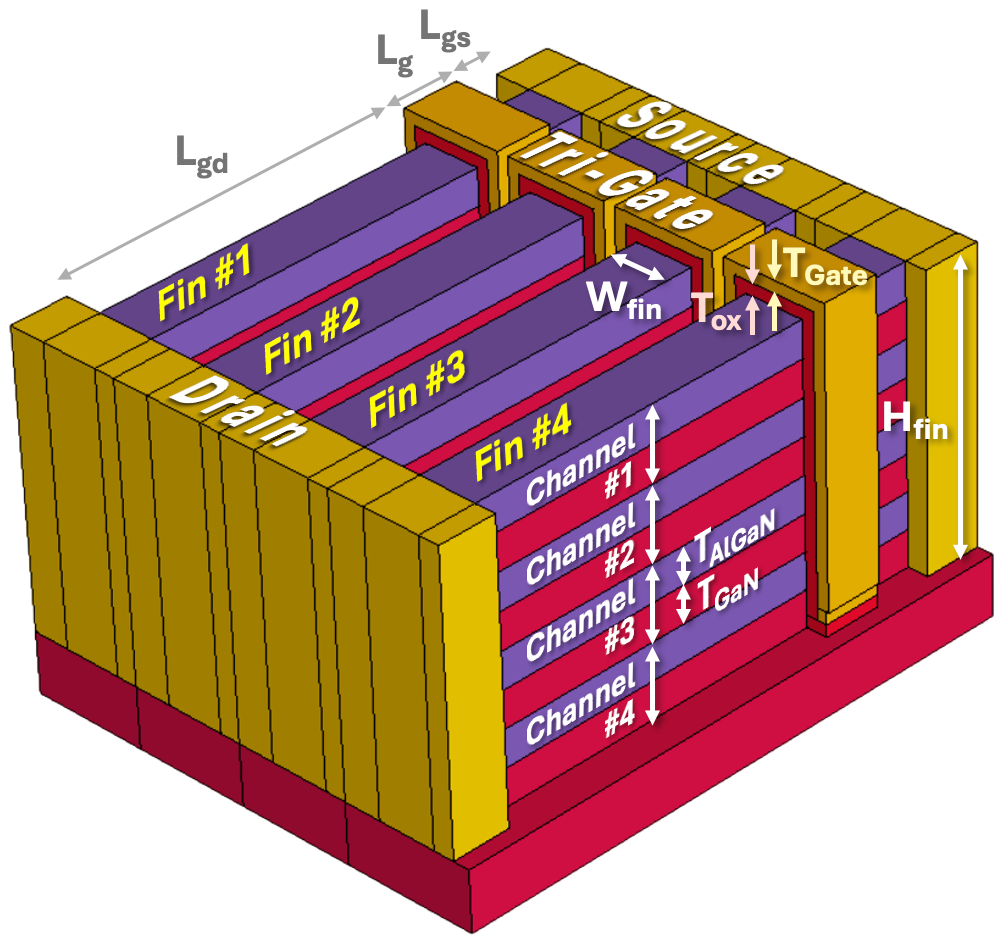}
    \end{minipage}\hfill
    \begin{minipage}[c]{0.24\textwidth}
        \centering
        \includegraphics[width=\linewidth]{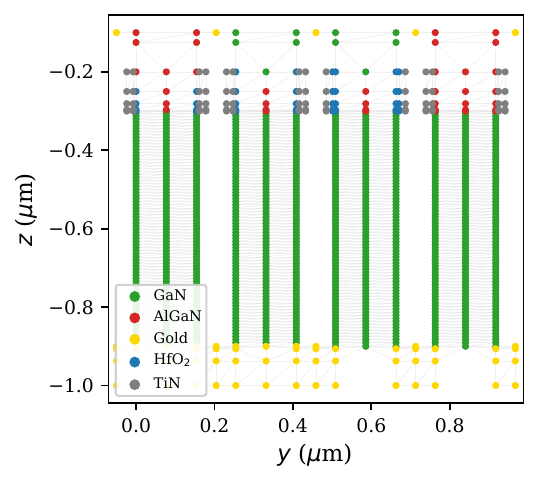}
    \end{minipage}
    \vspace{-7pt}
    \caption{Device parameters, geometry, and graph representation. \textbf{Left:}~Six design-space parameters and their exploration ranges. \textbf{Center:}~Physical structure showing the design variables. \textbf{Right:}~The same device represented as a mesh graph: nodes carry material identity, doping, and contact flags; edges carry Euclidean distance and a cross-material indicator.}
    \label{fig:finfet}
    \Description{Three-panel figure. Left: a table of multi-fin tri-gate FinFET design parameters with their symbols, baseline values, and exploration ranges. Center: a 3D rendering of the device with labelled geometric parameters. Right: a 2D cross-section of the corresponding mesh graph, with nodes colored by material (GaN, AlGaN, Gold, HfO2, TiN) and edges connecting adjacent mesh vertices.}
    \vspace{-10pt}
\end{figure*}

\subsection{Problem Formulation}
\label{sec:problem}
A multi-fin tri-gate gallium nitride (GaN) FinFET is parameterized by six continuous design variables bounded in Table~\ref{tab:ranges}:
\begin{equation*}
    \mathbf{x}=\big[\,T_{\mathrm{GaN}},\,T_{\mathrm{AlGaN}},\,W_{\mathrm{fin}},\,T_{\mathrm{ox}},\,T_{\mathrm{Gate}},\,N_{\mathrm{AlGaN}}\,\big]^\top\in\mathcal{X}\subset\mathbb{R}^6.
\end{equation*}
where $T_{\mathrm{GaN}}$ and $T_{\mathrm{AlGaN}}$ denote the thicknesses of the GaN channel and AlGaN barrier layers, respectively, $W_{\mathrm{fin}}$ is the fin width, $T_{\mathrm{ox}}$ is the oxide thickness, $T_{\mathrm{Gate}}$ is the gate thickness, and $N_{\mathrm{AlGaN}}$ is the AlGaN doping concentration.
The ranges are drawn from fabricated device dimensions in the literature and are broad---spanning up to two orders of magnitude for doping and up to $25\times$ for geometric dimensions---to avoid excluding non-obvious Pareto-optimal configurations; the active-learning loop of Sec.~\ref{sec:al} explores this space efficiently.
The fin count $N_{\mathrm{fins}}\!\in\!\{1,\dots,N_{\max}\}$ stratifies the design space; the same axis is reused for size extrapolation in Sec.~\ref{sec:experiments}. Fig.~\ref{fig:finfet} shows the device structure with annotated parameters (center) and how the same geometry maps to a mesh graph (right).

For a fixed $N_{\mathrm{fins}}$, the objective is the two-target maximization
\begin{equation}
    \max_{\mathbf{x}\in\mathcal{X}}\ \mathbf{f}(\mathbf{x};N_{\mathrm{fins}})=\big(I_{d,\max}(\mathbf{x}),V_{\mathrm{th}}(\mathbf{x})\big),\ \ V_{\mathrm{th}}>0,
    \label{eq:mo}
\end{equation}
with the enhancement-mode (E-mode) feasibility constraint $V_{\mathrm{th}}>0$ enforced as a post-hoc filter on the Pareto set. 
The ground-truth values $I_{d,\max}(\mathbf{x})$ and $V_{\mathrm{th}}(\mathbf{x})$ are produced by the Sentaurus Device (SDevice) drift-diffusion solver under a wall-time cap $\tau$; trials that time out or fail to converge are pruned from the training dataset.
While $\mathbf{f}$ is instantiated here with two standard device metrics for clarity, the training objective, mesh-graph message passing, and active-learning outer loop are \emph{agnostic to the choice and number of objectives and to the device family}. The GaN FinFET design and $(I_{d,\max},V_{\mathrm{th}})$ targets serve as a representative test case; the only device-specific component of the present implementation is the material-identity node feature (Sec.~\ref{sec:meshgraph}), a five-entry one-hot over \{GaN, AlGaN, Au, HfO\textsubscript{2}, TiN\}. Extension to arbitrary device families amounts to replacing this one-hot with a learned material embedding indexed by a general material library, a direction discussed in Sec.~\ref{sec:conclusion}.

\subsection{Mesh-Graph Representation}
\label{sec:meshgraph}
Surrogate input is the tetrahedral mesh produced by the Sentaurus Structure Editor (SDE) before any drift-diffusion solve is attempted. For each candidate design $\mathbf{x}$, a graph $G(\mathbf{x})\!=\!(\mathcal{V},\!\mathcal{E},\!\mathbf{X}_\mathcal{V},\!\mathbf{X}_\mathcal{E})$ is assembled once and re-used across all nine bias snapshots.

\paragraph{Vertices and edges.}
Each mesh vertex becomes a graph node; $|\mathcal{V}| = N$ ranges from a few thousand to tens of thousands nodes across the training set. Edges are extracted per tetrahedron as the $\binom{4}{2}=6$ unique vertex pairs per element, de-duplicated over the whole mesh, and stored bidirectionally so that every undirected edge $(i,j)$ contributes both $(i\!\rightarrow\!j)$ and $(j\!\rightarrow\!i)$ to the edge list. This bidirectional layout lets the finite-volume divergence of Sec.~\ref{sec:surrogate} be evaluated with a single scatter-add over the source index.

\paragraph{Node features.}
Each node $v\in\mathcal{V}$ carries a feature vector $\mathbf{x}_v\in\mathbb{R}^{17}$ grouped by physical origin (geometry $3$ + material $5$ + doping $1$ + Al fraction $1$ + contact flags $3$ + bias $2$ + permittivity $1$ + volume tag $1$ = $17$):
\begin{itemize}[leftmargin=*, itemsep=0pt, topsep=1pt]
    \item \emph{Geometry (3).} Per-axis normalized coordinates $(\tilde{x},\tilde{y},\tilde{z})$: $\tilde{x}$ is normalized by the total channel length $L_{\mathrm{ch}}$ 
    $\tilde{y}$ by one fin pitch ${W_{\mathrm{fin}}\!+\!S_{\mathrm{fin}}}$; $\tilde{z}$ by the per-sample $z$-bounding-box range. The $\tilde{y}$-by-pitch normalization enables size extrapolation by mapping each interior fin to a consistent $[0,1)$ local coordinate frame across ${N_{\mathrm{fins}}\!=\!1}$.
    \item \emph{Material identity (5).} One-hot encoding over the bulk material set ${\{\mathrm{GaN},\mathrm{AlGaN},\mathrm{Gold},\mathrm{HfO}_2,\mathrm{TiN}\}}$, derived from the Sentaurus region membership of the tetrahedra incident to $v$.
    \item \emph{Doping (1).} The signed log transform ${\mathrm{sgn}(N_v)\,\log_{10}(|N_v|+1)}$ of the local net doping concentration, which is approximately linear in volts and compresses the $10^{16}$--$10^{20}\,\mathrm{cm}^{-3}$ dynamic range without losing sign.
    \item \emph{Aluminum mole fraction (1).} The Al content of the AlGaN layer (Al\textsubscript{x}Ga\textsubscript{1-x}N), left in raw form.
    \item \emph{Contact flags (3).} Binary indicators for source ($S$), drain ($D$), and gate ($G$) membership. These also define the Dirichlet boundary set that is masked out of the data and physics losses in Sec.~\ref{sec:surrogate}.
    \item \emph{Applied bias (2).} Gate-source $V_{\mathrm{GS}}$ and drain-source $V_{\mathrm{DS}}$ biases broadcast to all nodes. Each Sentaurus run produces \emph{nine} graphs with sharing topology but different biases: five gate-sweep points at ${V_{\mathrm{DS}}=0}$ plus four drain-sweep points at ${V_{\mathrm{GS}}=V_{\mathrm{GS},\max}}$. These snapshots suffice to extract both $I_{d,\max}$ and $V_{\mathrm{th}}$ in post-processing.
    \item \emph{Relative permittivity (1).} A table lookup over the material identity, providing the surrogate with the local $\varepsilon_r$ that would otherwise have to be inferred from the one-hot encoding.
    \item \emph{Sparse volume tag (1).} A log-compressed Sentaurus volume field, retained for compatibility, not used as loss or uncertainty weight.
\end{itemize}

\paragraph{Edge features.}
Each directed edge $(i,j)\in\mathcal{E}$ carries $\mathbf{e}_{ij}\in\mathbb{R}^5$: the Euclidean distance $d_{ij}=\|\mathbf{p}_j-\mathbf{p}_i\|$, the displacement $\Delta\mathbf{p}_{ij}=\mathbf{p}_j-\mathbf{p}_i$, and a binary cross-material indicator $\mathbf{1}[\mathrm{mat}(i)\neq\mathrm{mat}(j)]$. The cross-material flag gives attention a cheap, physically meaningful way to separate interfacial edges (gate stack, AlGaN/GaN heterojunction) from intra-material bonds, which is where the current-continuity residual in Sec.~\ref{sec:surrogate} is sharpest.

\paragraph{Targets.}
Every node carries a three-dimensional target $\mathbf{y}_v=(\psi_v,\Phi_{n,v},\Phi_{p,v})\in\mathbb{R}^3$, where $\psi$ is the electrostatic potential and $\Phi_n$, $\Phi_p$ are the electron and hole electrochemical (quasi-Fermi) potentials produced by the Sentaurus drift-diffusion solver. These three scalar fields are the fundamental unknowns of the drift-diffusion system; all downstream figures of merit are recovered from them by post-processing (Sec.~\ref{sec:al}, \texttt{device\_metrics}).

\paragraph{Notation.}
Throughout the paper, $G_k(\mathbf{x})$ denotes the graph of design $\mathbf{x}$ at bias snapshot ${k\in\{1,\dots,9\}}$, and $f_\theta(G_k)\in\mathbb{R}^{N\times 3}$ denotes the per-node prediction of a surrogate with parameters $\theta$.

\subsection{Physics-Informed GAT Surrogate}
\label{sec:surrogate}

\paragraph{Architecture.}
The surrogate $f_\theta$ is a residual graph attention network operating on the mesh graph of Sec.~\ref{sec:meshgraph}. A two-layer multilayer perceptron (MLP) encoder maps each node-feature vector $\mathbf{x}_v$ to a $D$-dimensional embedding $\mathbf{h}^{(0)}_v$. A stack of $K$ message-passing blocks then updates the embedding according to
\begin{equation*}
    \mathbf{h}^{(k)}_v\;=\;\mathrm{LN}\!\big(\mathbf{h}^{(k-1)}_v + \mathbf{g}^{(k)}_v\big),
\end{equation*}
where $\mathbf{g}^{(k)}_v$ is the edge-conditioned GATv2 message~\cite{brody_iclr_2022} aggregated from the previous-layer neighborhood of $v$, using the edge feature vector $\mathbf{e}_{ij}$ as an attention bias. Each block uses four attention heads, an edge-feature projection of dimension $5$, and a LeakyReLU activation; the residual-plus-LayerNorm wrapper is essential for the eight-layer stack to train stably. A two-layer MLP decoder then maps $\mathbf{h}^{(K)}_v$ to the three-dimensional per-node prediction ${\hat{\mathbf{y}}_v=(\hat\psi_v,\hat\Phi_{n,v},\hat\Phi_{p,v})}$. The default configuration used throughout the paper is ${K\!=\!8}$ layers, hidden width ${D\!=\!128}$, four attention heads, node input dimension $17$, and edge input dimension $5$, for a total of $1{,}111{,}683$ trainable parameters. The network returns a per-node prediction: no global readout is used, so the same weights apply to meshes of any size, with inference bounded only by GPU memory (a dynamic-batching cap of ${\sim}100{,}000$ nodes per batch on an RTX~4090 in practice).

\paragraph{Why graph attention on the mesh.}
\textit{(i)}~GATv2 attention conditioned on edge features $\mathbf{e}_{ij}$ weights neighbors by physical attributes (distance, cross-material flag) rather than topology alone, matching the anisotropy of transport at heterojunctions. \textit{(ii)}~Eight message-passing hops cover the channel-to-contact diameter of these meshes; the residual-plus-LayerNorm wrapper stabilizes the stack at hidden width $128$~\cite{brandstetter_iclr_2022}. \textit{(iii)}~No global readout is used, so the same weights apply unchanged to larger meshes.

\paragraph{Training objective.}
The loss combines an unweighted per-node data term on the three predicted fields with a physics term that enforces steady-state current continuity on the same mesh edges:
\begin{equation}
    \mathcal{L}(\theta)\;=\;w_{\mathrm{data}}\,\mathcal{L}_{\mathrm{data}}(\theta) \;+\; \lambda_{\mathrm{phys}}(t)\,w_{\mathrm{cont}}\,\mathcal{L}_{\mathrm{cont}}(\theta),
    \label{eq:loss-total}
\end{equation}
with fixed base weights $w_{\mathrm{data}}\!=\!1$ and $w_{\mathrm{cont}}\!=\!0.1$, and a time-dependent physics weight $\lambda_{\mathrm{phys}}(t)$ described below.

\paragraph{Data loss.}
Targets and predictions are mapped to per-field $z$-score space using per-field means and standard deviations computed once at training start over \emph{interior} (non-contact) nodes of the training set. Let $\mathcal{V}_{\mathrm{int}}$ denote the set of nodes with no source, drain, or gate contact flag. The data loss is the unweighted mean squared error
\begin{equation}
    \mathcal{L}_{\mathrm{data}}\;=\;\frac{1}{|\mathcal{V}_{\mathrm{int}}|}\sum_{v\in\mathcal{V}_{\mathrm{int}}}\big\|\tilde{\mathbf{y}}_v - \tilde{f}_\theta(G)_v\big\|_2^2,
    \label{eq:data-loss}
\end{equation}
where $\tilde{(\cdot)}$ denotes $z$-score normalization. Contact nodes are excluded because $\Phi_n$ and $\Phi_p$ at source, drain, and gate contacts are Dirichlet-pinned to the applied bias, so the mean squared error on those nodes is small and would bleed gradient signal away from the channel region where the physics is nontrivial. The same interior-mean formulation is used for the uncertainty definition in Sec.~\ref{sec:al}.

\paragraph{Current-continuity residual.}
The physics term enforces ${\nabla\!\cdot\!\mathbf{J}_n=0}$ and ${\nabla\!\cdot\!\mathbf{J}_p=0}$ at interior nodes in a finite-volume sense. Given the predicted fields, carrier densities are reconstructed from Boltzmann statistics,
\begin{equation}
    n_v\,=\,N_C\,\exp\!\Big(\tfrac{\psi_v-\Phi_{n,v}}{V_T}\Big),\quad
    p_v\,=\,N_V\,\exp\!\Big(\tfrac{\Phi_{p,v}-\psi_v}{V_T}\Big),
    \label{eq:boltz}
\end{equation}
with $V_T\!=\!k_B T/q\!\approx\!25.85\,\mathrm{mV}$ at ${T\!=\!300\,\mathrm{K}}$ and the effective densities of states ${N_C\!=\!2.23{\times}10^{18}\,\mathrm{cm}^{-3}}$, ${N_V\!=\!4.62{\times}10^{19}\,\mathrm{cm}^{-3}}$ for GaN as in the Sentaurus material models. The arguments of the exponentials are clamped to $[-30,30]$ to keep the forward pass inside float32 range. An Einstein-relation edge flux follows as
\begin{equation}
    \tilde{J}^{n}_{ij}\;=\;\tfrac{1}{2}(n_i+n_j)\,\frac{\Phi_{n,j}-\Phi_{n,i}}{d_{ij}}\,\cdot\,\frac{1}{J_{\mathrm{ref}}},
    \label{eq:edge-flux}
\end{equation}
and analogously for holes, with ${J_{\mathrm{ref}}=N_C V_T}$ a purely numerical normalization that keeps squared fluxes inside float32. Since the edge list is bidirectional and the quasi-Fermi difference $\Phi_{n,j}-\Phi_{n,i}$ flips sign between $(i,j)$ and $(j,i)$, the finite-volume divergence at node $v$ is obtained exactly by accumulating signed edge fluxes with a single scatter-add over the source index,
\begin{equation*}
    (\nabla\!\cdot\!\tilde{\mathbf{J}}_n)_v\;=\;\sum_{j:(v,j)\in\mathcal{E}} \tilde{J}^{n}_{vj},
\end{equation*}
with no double-counting. The continuity residual is then
\begin{align}
    \mathcal{L}_{\mathrm{cont}} \;=\; \frac{1}{|\mathcal{V}_{\mathrm{int}}|}\sum_{v\in\mathcal{V}_{\mathrm{int}}}\Big[& \log\!\big(1+(\nabla\!\cdot\!\tilde{\mathbf{J}}_n)_v^{2}\big) 
    + \log\!\big(1+(\nabla\!\cdot\!\tilde{\mathbf{J}}_p)_v^{2}\big)\Big].
    \label{eq:cont-loss}
\end{align}
The $\log(1+x^2)$ form is a robust quadratic penalty that bounds the gradient contribution from outliers without hard clipping; a $|x|\!<\!10^{18}$ safeguard keeps $x^2$ representable in float32. Contact nodes are masked out since current legitimately enters and exits there. Relative to the solver's Scharfetter--Gummel box scheme, the residual uses edge lengths with unit face areas, edge-averaged Boltzmann densities with constant mobility, and no generation--recombination: it is a soft regularizer, not a re-implementation of the solver.

\paragraph{Physics-weight ramp.}
The physics weight is ramped from zero:
\begin{equation*}
    \lambda_{\mathrm{phys}}(t)\;=\;\begin{cases}
        0 & t<t_{\mathrm{start}} \\
        \lambda_{\max}\,\cdot\,\frac{t-t_{\mathrm{start}}}{t_{\mathrm{ramp}}} & t_{\mathrm{start}}\leq t<t_{\mathrm{start}}+t_{\mathrm{ramp}} \\
        \lambda_{\max} & \text{otherwise}
    \end{cases}
\end{equation*}
with ${t_{\mathrm{start}}=20}$, ${t_{\mathrm{ramp}}=150}$ epochs, and ${\lambda_{\max}=0.1}$. The ramp is motivated by the practical observation that a cold start with ${\lambda_{\mathrm{phys}}>0}$ mixes two non-trivial gradient fields before either has a low-loss attractor; data fitting is therefore allowed to establish a sensible field before the continuity residual begins to shape it. The value ${\lambda_{\max}=0.1}$ is a design choice consistent with the data and continuity terms being of comparable scale after $z$-score normalization; a systematic sweep is left for future work.

\paragraph{Training setup.}
Optimization uses AdamW with base learning rate ${10^{-3}}$ and weight decay ${10^{-4}}$, a cosine-annealing schedule over $500$ epochs, and gradient clipping at $5$. Batching is dynamic: instead of a fixed batch size, graphs are grouped until their total node count reaches a cap of roughly $28{,}000$ nodes. A fixed batch size of eight graphs easily runs out of memory on the $N_{\mathrm{fins}}\!\geq\!3$ tail of the training distribution (largest graph ${\sim}10.7$k nodes, median ${\sim}3.8$k); dynamic batching keeps throughput high while respecting this mesh-size imbalance. A single training epoch takes approximately \emph{one minute} on an RTX~4090 under the dynamic-batching cap used throughout (${\sim}8$\,h for the full 500-epoch schedule).

\subsection{TCAD-in-the-Loop Active Learning}
\label{sec:al}
The surrogate is embedded in an outer loop in which a deep ensemble of $M$ independently seeded models selects informative designs for full Sentaurus simulation. The use of the SDE for mesh generation rather than the full drift-diffusion solve for the candidate pool is the key speed lever: a candidate costs a mesh (${\sim}30\,\mathrm{s}$), not a solution (${\sim}20$--${60}$ minutes).

\paragraph{Ensemble.}
The loop maintains $M=5$ independently seeded copies $\{f_{\theta_j}\}_{j=1}^{M}$ of the \mbox{GATv2} surrogate, each trained from scratch on the current dataset. Five members offer a reasonable variance estimate while staying within the training budget of a two-GPU workstation.

\paragraph{Mesh-only candidate generation.}
Candidates are sampled uniformly over ${N_{\mathrm{fins}}\in\{1,\dots,N_{\max}\}}$, with the six continuous variables drawn per-stratum from the ranges in Table~\ref{tab:ranges}. The fin-spacing constraint ${T_{\mathrm{ox}}+T_{\mathrm{Gate}}<S_{\mathrm{fin}}/2}$ is enforced by rejection. For each accepted tuple, only the Sentaurus SDE (mesh generation) step is executed, producing the full set of nine bias-snapshot graphs ${\{G_k(\mathbf{x})\}_{k=1}^{9}}$ with target fields left as zeros. These zero-target graphs are never used for training or evaluation; they only carry the mesh topology and node/edge features that the surrogate needs for inference.

\paragraph{Per-candidate uncertainty.}
For a candidate $\mathbf{x}$ and snapshot $k$, let $f_{\theta_j}(G_k)_v\in\mathbb{R}^3$ denote the per-node prediction of ensemble member $j$ at node $v$, and let ${\bar f(G_k)_v=\tfrac{1}{M}\sum_{j}f_{\theta_j}(G_k)_v}$ be the ensemble mean. The per-node ensemble variance is
\begin{equation*}
    \sigma^2_v(\mathbf{x},k)\;=\;\frac{1}{M}\sum_{j=1}^{M}\big\|f_{\theta_j}(G_k)_v\,-\,\bar f(G_k)_v\big\|_2^2.
\end{equation*}
The per-snapshot scalar uncertainty is obtained by averaging $\sigma^2_v$ over \emph{interior} nodes only, using the same mask as the data loss in Sec.~\ref{sec:surrogate}, and taking a square root:
\begin{equation*}
    u_k(\mathbf{x})\;=\;\sqrt{\frac{1}{|\mathcal{V}_{\mathrm{int}}|}\sum_{v\in\mathcal{V}_{\mathrm{int}}}\sigma^2_v(\mathbf{x},k)}.
\end{equation*}
The candidate-level uncertainty is the worst-case over the nine bias snapshots,
\begin{equation}
    U(\mathbf{x})\;=\;\max_{k\in\{1,\dots,9\}}\;u_k(\mathbf{x}).
    \label{eq:u-cand}
\end{equation}

\paragraph{Predicted objectives.}
The ensemble-mean fields on the nine snapshots are passed through physics-based post-processing to obtain two scalar predictions $\hat I_{d,\max}(\mathbf{x})$ and $\hat V_{\mathrm{th}}(\mathbf{x})$. The drain current is the electron-plus-hole flux integrated over drain-boundary edges of the high-bias snapshot, normalized by the effective device width~$W_{\mathrm{eff}}$,
\begin{equation*}
    W_{\mathrm{eff}} = N_{\mathrm{fins}}\,(P + W_{\mathrm{fin}} + S_{\mathrm{fin}}),
\end{equation*}
where $P = 2\,H_{\mathrm{fin}}$ is the fin cross-sectional perimeter with $H_{\mathrm{fin}} = T_{\mathrm{AlGaN}} + T_{\mathrm{GaN}}$. The threshold voltage $\hat V_{\mathrm{th}}$ is the gate bias at which the mean electron density over the GaN channel mask crosses $10^{16}\,\mathrm{cm}^{-3}$, obtained by linear interpolation of the five gate-sweep snapshots at $V_{\mathrm{DS}}=0$.

\paragraph{Acquisition score.}
For the candidate pool $\mathcal{C}_t$ at iteration $t$, the uncertainty and each predicted objective are $z$-scored independently across $\mathcal{C}_t$ and combined in an upper-confidence-bound (UCB)-style score,
\begin{equation}
    s(\mathbf{x})\;=\;\alpha\,\big[z\!\big(\hat I_{d,\max}(\mathbf{x})\big)+z\!\big(\hat V_{\mathrm{th}}(\mathbf{x})\big)\big] \;+\;\beta\,z\!\big(U(\mathbf{x})\big),
    \label{eq:al-score}
\end{equation}
with $\alpha=0.5$ and $\beta=1.0$. The predicted-objective term captures the ``exploitation'' bias toward promising regions of design space, while the $U(\mathbf{x})$ term drives ``exploration'' into regions where the ensemble disagrees. The chosen weighting leans more on $U(\mathbf{x})$ than on the predicted objective: inference on a mesh candidate is orders of magnitude faster than a Sentaurus run, so the marginal cost of occasionally sampling a redundant informative candidate is smaller than the cost of missing an informative one.

\paragraph{Diversity filter.}
A greedy top-$K$ selection by $s(\mathbf{x})$ is applied, rejecting a candidate whose $L_2$ distance in the six-dimensional normalized parameter space falls below a threshold $\delta$ from any already-selected candidate \emph{with the same $N_{\mathrm{fins}}$}. Cross-stratum rejection is deliberately disabled so that single-fin and multi-fin candidates never exclude each other, preserving coverage of every $N_{\mathrm{fins}}$ stratum.

\paragraph{Plateau stopping criterion.}
The loop tracks the median candidate-level uncertainty
 ${\bar U_t\;=\;\mathrm{median}_{\mathbf{x}\in\mathcal{C}_t^{\mathrm{ok}}}\,U(\mathbf{x})
}$
over successfully scored candidates of iteration $t$. 

The relative improvement $\rho_t=(\bar U_{t-1}-\bar U_t)/\max(\bar U_{t-1},\varepsilon_0)$ is monitored, and the loop terminates when ${\rho_t<\varepsilon}$ for $\ell$ consecutive iterations or when the Sentaurus budget $N$ is exhausted. Defaults: ${\varepsilon=0.02}$, ${\ell=3}$. 

\begin{algorithm}[!t]
    \SetAlFnt{\footnotesize}
    \DontPrintSemicolon
    \SetAlgoLined
    \caption{TCAD-in-the-loop active learning with a mesh-native GAT ensemble}
    \label{alg:alpha_explore}
    \KwIn{design space $\mathcal{X}\subset\mathbb{R}^6$, strata count $N_{\max}$, bootstrap size $n_0$, Sentaurus budget $N$, ensemble size $M{=}5$, candidate pool size $n_c$, batch size $b$, acquisition weights $\alpha{=}0.5,\beta{=}1.0$, diversity threshold $\delta$, plateau tolerance $\varepsilon$, patience $\ell$, wall-time cap $\tau$}
    \KwOut{Sentaurus-simulated dataset $\mathcal{D}_T$}
    Draw $n_0$ samples stratified over $N_{\mathrm{fins}}\!\in\!\{1,\dots,N_{\max}\}$; run Sentaurus (SDE+SDevice) with cap $\tau$; form $\mathcal{D}_0$ (prune timeouts/failures).\;
    \For{$t=1,2,\dots$}{
        Train $\{f_{\theta_j}\}_{j=1}^{M}$ on $\mathcal{D}_{t-1}$ (five independently seeded members).\;
        Build mesh-only pool $\mathcal{C}_t\subset\mathcal{X}$ of size $n_c$: draw params stratified over $N_{\mathrm{fins}}$, for $N_{\mathrm{fins}}\!>\!1$ reject those violating $T_{\mathrm{ox}}+T_{\mathrm{Gate}}<S_{\mathrm{fin}}/2$, run SDE only to obtain nine bias-snapshot graphs per candidate.\;
        \For{$\mathbf{x}\in\mathcal{C}_t$ \textnormal{\textbf{in parallel}}}{
            \For{snapshot $k=1,\dots,9$}{
                predict $f_{\theta_j}(G_k(\mathbf{x}))$ for $j=1,\dots,M$\;
                compute per-node variance $\sigma^2_v(\mathbf{x},k)$ and interior-masked scalar $u_k(\mathbf{x})$\;
            }
            $U(\mathbf{x})\leftarrow\max_k u_k(\mathbf{x})$ \hfill \eqref{eq:u-cand}\;
            $(\hat I_{d,\max},\hat V_{\mathrm{th}})\leftarrow\texttt{device\_metrics}\big(\bar f(G_{1..9}(\mathbf{x}))\big)$\;
        }
        $z$-score $U$, $\hat I_{d,\max}$, and $\hat V_{\mathrm{th}}$ independently across $\mathcal{C}_t$; score $s(\mathbf{x})$ by~\eqref{eq:al-score}.\;
        Select top-$b$ by $s$ with same-$N_{\mathrm{fins}}$ $\delta$-diversity filter in normalized parameter space.\;
        Dispatch selections to Sentaurus (SDE+SDevice, cap $\tau$); ingest successes into $\mathcal{D}_t$.\;
        $\bar U_t\leftarrow\mathrm{median}_{\mathbf{x}\in\mathcal{C}_t^{\mathrm{ok}}} U(\mathbf{x})$;\quad
        $\rho_t\leftarrow(\bar U_{t-1}-\bar U_t)/\max(\bar U_{t-1},\varepsilon_0)$.\;
        \If{$\rho_t<\varepsilon$ \textnormal{for} $\ell$ \textnormal{consecutive iterations} \textbf{or} $|\mathcal{D}_t|\ge N$}{
            \textbf{break}
        }
    }
    \Return $\mathcal{D}_T$.\;
\end{algorithm}

\section{Experiments}
\label{sec:experiments}
This section evaluates the mesh-native physics-informed GAT surrogate of Sec.~\ref{sec:method} on three questions: \textit{(i)}~does it reconstruct the drift-diffusion unknowns accurately, and does accuracy degrade gracefully at fin counts never seen during training? \textit{(ii)}~how does its inference cost compare against the full Sentaurus drift-diffusion solve it substitutes inside the active-learning loop? \textit{(iii)}~does per-stratum Pareto post-processing of the dataset assembled by the loop surface useful E-mode designs?

\subsection{Setup}
\label{sec:exp-setup}
\paragraph{Dataset.}
The training set consists of $1{,}786$ bias snapshots from $200$ Sentaurus Device runs (device geometries) \emph{assembled over the course of the active-learning loop of Sec.~\ref{sec:al}}: $147$ bootstrap designs generated by uniform sampling over Table~\ref{tab:ranges} seed the ensemble, and subsequent AL iterations contribute $53$ further designs selected by the acquisition score of Eq.~\eqref{eq:al-score}; the loop concentrated these where the surrogate was least certain ($64\%$ of the selected designs have $N_{\mathrm{fins}}\!\geq\!3$, vs.\ $10\%$ of the bootstrap). The set spans $N_{\mathrm{fins}}\!\in\!\{1,\dots,5\}$; a disjoint validation set of $1{,}017$ snapshots ($113$ designs) across the same fin counts is used for model selection. A data-scaling ablation indicates that the surrogate is sample-efficient: the in-distribution ranking fidelity of the predicted $I_{d,\max}$ against TCAD on held-out designs saturates (Spearman $\rho\!\approx\!0.99$) by ${\sim}25$ training designs. Each ensemble member is trained for $500$ epochs (${\sim}8$\,h on an RTX~4090 at ${\sim}1$\,min/epoch). To evaluate generalization beyond the training range, a separate benchmark test set is generated by running the full Sentaurus pipeline at $N_{\mathrm{fins}}\!=\!1$ through~$40$ with five independent random parameter samples per fin count; the $N_{\mathrm{fins}}\!>\!5$ strata are never seen during training and measure out-of-distribution extrapolation under the per-fin-pitch normalization of Sec.~\ref{sec:meshgraph}.

\paragraph{Ensemble.}
The reported numbers use a deep ensemble of $M{=}5$ independently seeded GATv2 members, trained from scratch with the physics-informed loss of~\eqref{eq:loss-total}.

\paragraph{Hardware.}
All training and inference run on a single workstation with two NVIDIA RTX~4090 GPUs, an AMD Ryzen~9 7950X, and 124\,GB of RAM. Sentaurus Device wall times in Sec.~\ref{sec:exp-runtime} are collected from the same workstation under realistic license-bounded worker concurrency.

\paragraph{Metrics.}
Prediction quality is reported as the unweighted RMSE (in volts) of each drift-diffusion field $\psi$, $\Phi_n$, $\Phi_p$ averaged over mesh nodes and over samples in each $N_{\mathrm{fins}}$ stratum. RMSE rather than relative $L_2$ is used because $\Phi_n,\Phi_p$ collapse to zero over buffer and oxide regions, making a relative-$L_2$ denominator numerically unstable.

\begin{figure*}[t]
    \centering
    \vspace{-10pt}
    \includegraphics[width=\linewidth]{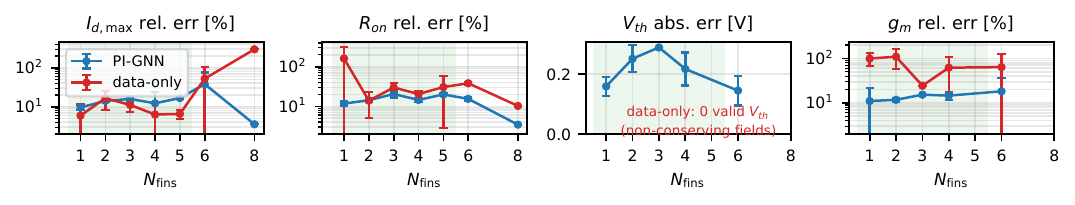}
    \Description{Four-panel line plot of device-metric error versus fin count from 1 to 8 for two models, PI-GNN (physics-informed) and a data-only ablation. Panels show relative error of I_d,max, R_on, and g_m on a log scale and absolute error of V_th in volts; markers are medians per fin count with standard-error bars. A green-shaded band marks the training range Nfins 1-5. The data-only curve diverges beyond the training range on I_d,max and is absent from the V_th panel, annotated as zero valid V_th, while PI-GNN stays bounded in every panel.}
    \vspace{-10pt}
    \caption{Device-metric accuracy vs.\ $N_{\mathrm{fins}}$ on held-out designs for the physics-informed surrogate (PI-GNN) and a data-only ablation (same architecture, $\lambda_{\max}\!=\!0$). FoMs are read back by uploading the predicted $(\psi,\Phi_n,\Phi_p)$ fields into Sentaurus Device (no re-solve), so ground truth and surrogate share one extraction route. Panels report the relative error of $I_{d,\max}$, $R_{\mathrm{on}}$, and $g_m$ (log scale) and the absolute error of $V_{\mathrm{th}}$; markers are per-fin-count medians with standard-error bars. The shaded region indicates the training range $N_{\mathrm{fins}}\in\{1,\ldots,5\}$. The data-only fields yield no valid $V_{\mathrm{th}}$ (non-conserving fields) and diverge on $I_{d,\max}$ beyond the training range.}
    \label{fig:scalability}
    \vspace{-12pt}
\end{figure*}


\subsection{Size Extrapolation on Device Metrics}
\label{sec:exp-scalability}
Figure~\ref{fig:scalability} reports the device-metric error on the four standard FoMs---drain current $I_{d,\max}$, on-resistance $R_{\mathrm{on}}$, threshold voltage $V_{\mathrm{th}}$, and peak transconductance $g_m$---as a function of $N_{\mathrm{fins}}$ on $39$ held-out designs with $1$--$8$ fins, for the physics-informed surrogate (PI-GNN) and a data-only ablation (same architecture, $\lambda_{\max}\!=\!0$). Ground-truth and surrogate FoMs are extracted by the same route: the predicted $(\psi,\Phi_n,\Phi_p)$ fields are uploaded into Sentaurus Device, which reads the FoMs from them exactly as from its own solution (no re-solve); the closed-form post-processing of Sec.~\ref{sec:al} remains the fast in-loop estimator. $V_{\mathrm{th}}$ is reported as an absolute error because the true threshold crosses zero across the design space. Within the training range the two surrogates are comparable on $I_{d,\max}$ (median error $14\%$ for PI-GNN vs.\ $7\%$ for data-only), while PI-GNN is already more accurate on $R_{\mathrm{on}}$ ($16\%$ vs.\ $24\%$) and $g_m$ ($14\%$ vs.\ $68\%$). Beyond the training range the data-only ablation diverges: its $I_{d,\max}$ error grows to $53\%$ at $N_{\mathrm{fins}}\!=\!6$ and ${\approx}300\%$ on the single $N_{\mathrm{fins}}\!=\!8$ design, whereas PI-GNN stays bounded ($38\%$ and $4\%$). The data-only fields yield \emph{no} valid $V_{\mathrm{th}}$ on any design (they do not conserve current, so the reconstructed transfer characteristic is unusable), while PI-GNN's $V_{\mathrm{th}}$ error stays within $0.14$--$0.29$\,V. The physics loss thus acts as a regularizer that keeps the derived device metrics usable under extrapolation, porting the learned stencil to unseen fin counts under the per-fin-pitch normalization of Sec.~\ref{sec:meshgraph}.

\paragraph{Per-field RMSE.}
At the field level, PI-GNN reconstructs $\psi$, $\Phi_n$, and $\Phi_p$ with per-field RMSE of $0.16$--$1.3$\,V in the training range, growing by ${\approx}2$--$12\times$ at $N_{\mathrm{fins}}\!=\!38$; the data-only ablation starts tighter (${\approx}0.1$\,V per field in-distribution) but grows ${\approx}18$--$22\times$ over the same range. The field RMSE is computed over all mesh nodes, whereas FoMs are extracted from electrically dominant regions and cross-bias post-processing, so local field errors in weakly relevant regions inflate RMSE without proportionally degrading the derived metrics; this is why PI-GNN's larger in-distribution field RMSE does not translate into weaker device-metric accuracy in Fig.~\ref{fig:scalability}.

\begin{figure}[t]
    \centering
    \includegraphics[width=0.64\linewidth]{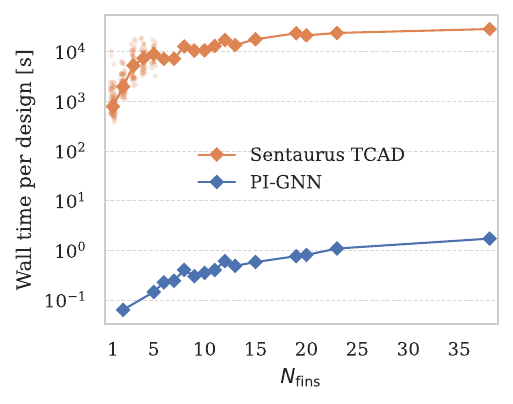}
    \Description{Log-scale line plot comparing Sentaurus TCAD and PI-GNN per-design wall time vs fin count, showing a roughly four-order-of-magnitude gap.}
    \vspace{-8pt}
    \caption{Per-design wall time vs.\ $N_{\mathrm{fins}}$ (log scale) for Sentaurus Device (drift--diffusion solve) and PI-GNN on the same meshes. Shaded bands show the 5th--95th percentile interval.}
    \label{fig:runtime}
    \vspace{-10pt}
\end{figure}

\subsection{Simulator vs.~Surrogate Runtime}
\label{sec:exp-runtime}
Figure~\ref{fig:runtime} contrasts the wall time of a full Sentaurus Device drift-diffusion solve against a single GAT forward pass on the same mesh, both as a function of $N_{\mathrm{fins}}$. Sentaurus Device wall time grows super-linearly with fin count---from roughly $19$\,min at $N_{\mathrm{fins}}\!=\!2$ to over $7.5$\,h at $N_{\mathrm{fins}}\!=\!38$---as the mesh and Newton iteration count increase; a single GAT forward pass (graph construction from cached mesh plus one ensemble-member inference on an RTX~4090) grows roughly linearly from $65$\,ms to $1.7$\,s over the same range. The resulting speedup is consistently \emph{four orders of magnitude} (${\approx}1.6{\times}10^4$--$4.1{\times}10^4{\times}$) and is the lever that makes the active-learning loop of Sec.~\ref{sec:al} worth implementing: an uncertainty-guided acquisition score only pays off if evaluating the score on a large candidate pool is cheap relative to a TCAD run.

\subsection{Pareto Exploitation and Design Selection}
\label{sec:exp-pareto}
Once the active-learning loop terminates, the simulated dataset $\mathcal{D}_T$ is treated as the candidate pool for Pareto post-processing. The E-mode feasibility filter of~\eqref{eq:mo} removes configurations with $V_{\mathrm{th}}\!\leq\!0$, and a per-stratum non-dominated sort produces a two-target front in the $(I_{d,\max},V_{\mathrm{th}})$ plane separately for each fin count. A farthest-first diversity selector then spreads the recommended design points across the feasible subspace rather than clustering them at a single Pareto knee.

\paragraph{Summary.}
The physics-informed GAT surrogate reconstructs the three drift-diffusion fields with per-field RMSE of $0.16$--$1.3$\,V in the training range and degrades more gracefully under extrapolation than a data-only ablation, confirming the regularization benefit of the physics loss. Inference runs \emph{four orders of magnitude faster} than the Sentaurus solve it replaces, and Pareto post-processing exposes the design trade-offs of multi-fin GaN FinFETs in a single pass over the simulated dataset. These results rest on a training set assembled by the active-learning loop of Sec.~\ref{sec:al}, whose uncertainty-driven acquisition keeps the training-data cost practical.

\section{Conclusion}
\label{sec:conclusion}
This paper presents a mesh-native physics-informed graph attention surrogate for drift-diffusion simulation of multi-fin GaN FinFETs, targeting global design-space exploration rather than local calibration around a known baseline. The model operates directly on the tetrahedral mesh produced by the Sentaurus mesh generator, predicts the three drift-diffusion unknowns $(\psi,\Phi_n,\Phi_p)$ at every node, and is trained with a current-continuity residual loss that mirrors the finite-volume discretization of the TCAD solver. A deep ensemble of GATv2 members provides per-node uncertainty that drives a TCAD-in-the-loop active-learning loop: the candidate pool requires only mesh generation, and the expensive drift-diffusion solve is reserved for acquisition-selected designs.

On a benchmark test set spanning $N_{\mathrm{fins}}\!=\!1$ through~$40$, the surrogate reconstructs all three fields with sub-volt per-field RMSE that degrades gracefully under $5\!\to\!40$ fin-count extrapolation, confirming that the per-fin-pitch coordinate normalization ports the learned stencil to larger arrays. An ablation confirms that the physics loss acts as a regularizer, trading in-distribution accuracy for a flatter extrapolation error curve. Inference takes $65$\,ms to $1.7$\,s per mesh on a single RTX~4090 depending on fin count, roughly \emph{four orders of magnitude} below the Sentaurus wall time, making screening of $10^3$--$10^4$ mesh-only candidates per active-learning round practical. The surrogate's advantage over black-box optimizers rests on four compounding inductive biases: the mesh graph carries device geometry, the finite-volume residual enforces carrier-transport physics, the deep ensemble surfaces epistemic uncertainty, and the active-learning loop drives that uncertainty down preferentially near the Pareto frontier.


\paragraph{Scope.}
This study trains on designs with $N_{\mathrm{fins}} \leq 5$ and uses a node-feature encoding tailored to the GaN/AlGaN/HfO$_2$/TiN/Au stack. The mesh-native formulation and physics-informed training are not tied to these choices, and extend naturally to larger device sizes and alternative material stacks.

\paragraph{Future work.}
Three directions are immediate. \textit{(i)}~\emph{Closing the inverse-design loop}: coupling the predicted fields and per-node uncertainty to gradient-based optimization in design-parameter space, validated end-to-end against TCAD. \textit{(ii)}~\emph{Device-agnostic extension}: replacing the fixed five-material one-hot with a learned material embedding indexed against a general material library, enabling transfer to other transistor topologies and to non-transistor TCAD problems that share the per-node field structure. \textit{(iii)}~Broadening the training range directly to cover higher fin counts and richer bias sweeps, and weighting the loss toward the gate stack and barrier, where the residual error concentrates.

\begin{acks}
This work was supported by the Center for Heterogeneous Integration of Micro Electronic Systems (CHIMES), one of seven centers in the Joint University Microelectronics Program (JUMP)~2.0, a Semiconductor Research Corporation (SRC) program sponsored by the Defense Advanced Research Projects Agency (DARPA), under Grant 2023-JU-3136. The authors thank Prof.~Debjit Pal for providing the distributed simulation infrastructure used in this work.
\end{acks}

\clearpage
\balance
\bibliographystyle{ACM-Reference-Format}
\bibliography{references}

\end{document}